\documentclass[10pt,twocolumn,letterpaper]{article}

\usepackage{wacv}
\usepackage{times}
\usepackage{epsfig}
\usepackage{graphicx}
\usepackage{amsmath}
\usepackage{amssymb}

\usepackage{makecell}
\usepackage{enumitem}
\usepackage{subfigure}

\usepackage[pagebackref=true,breaklinks=true,letterpaper=true,colorlinks,bookmarks=false]{hyperref}

\wacvfinalcopy 

\def\wacvPaperID{331} 

\ifwacvfinal\fi
\begin{document}

\title{Composition-Aware Image Aesthetics Assessment}

\author{
  Dong Liu, Rohit Puri, Nagendra Kamath, Subhabrata Bhattacharya\\
    Netflix Inc., Los Gatos, CA, USA\\
        {\tt\small \{dongl,rpuri,nkamath,sbhattacharya\}@netflix.com}
        }

\maketitle

\begin{abstract}
Automatic image aesthetics assessment is important for a wide variety of applications such as on-line photo suggestion, photo album management and image retrieval. Previous methods have focused on mapping the holistic image content to a high or low aesthetics rating. However, the composition information of an image characterizes the harmony of its visual elements according to the principles of art, and provides richer information for learning aesthetics. In this work, we propose to model the image composition information as the mutual dependency of its local regions, and design a novel architecture to leverage such information to boost the performance of aesthetics assessment. To achieve this, we densely partition an image into local regions and compute aesthetics-preserving features over the regions to characterize the aesthetics properties of image content. With the feature representation of local regions, we build a region composition graph in which each node denotes one region and any two nodes are connected by an edge weighted by the similarity of the region features. We perform reasoning on this graph via graph convolution, in which the activation of each node is determined by its highly correlated neighbors. Our method naturally uncovers the mutual dependency of local regions in the network training procedure, and achieves the state-of-the-art performance on the benchmark visual aesthetics datasets.                                      
\end{abstract}
\vspace{-6mm}

\section{Introduction}\label{sec:introduction}

Automatic aesthetics assessment is typically cast into a classification or regression problem where the image content is mapped to the aesthetics ratings provided by human annotators~\cite{Kong:ECCV16, Lu:MM14}. Early efforts focused on designing some hand-crafted aesthetics visual features and feeding them into conventional machine learning models~\cite{Datta:ECCV06, Dhar:CVPR11,  Marchesotti:IJCV15, Marchesotti:ICCV11, Ordonez:CVPR11, Tang:TMM13}. 
With the advance of deep Convolution Neural Networks (CNN)~\cite{Krizhevsky:NIPS12}, the recent works proposed to learn end-to-end deep models that jointly embed image content to visual features and infer aesthetics ratings, yielding the state-of-the-art performance on the benchmark image aesthetics assessment dataset~\cite{Jin:AAAI18, Lu:ICCV15, Murray:arxiv2017, Mai:CVPR16, Wang:ICCV17, Wang:PAMI18}. 

\begin{figure}
\begin{center}
\includegraphics[width=1\linewidth]{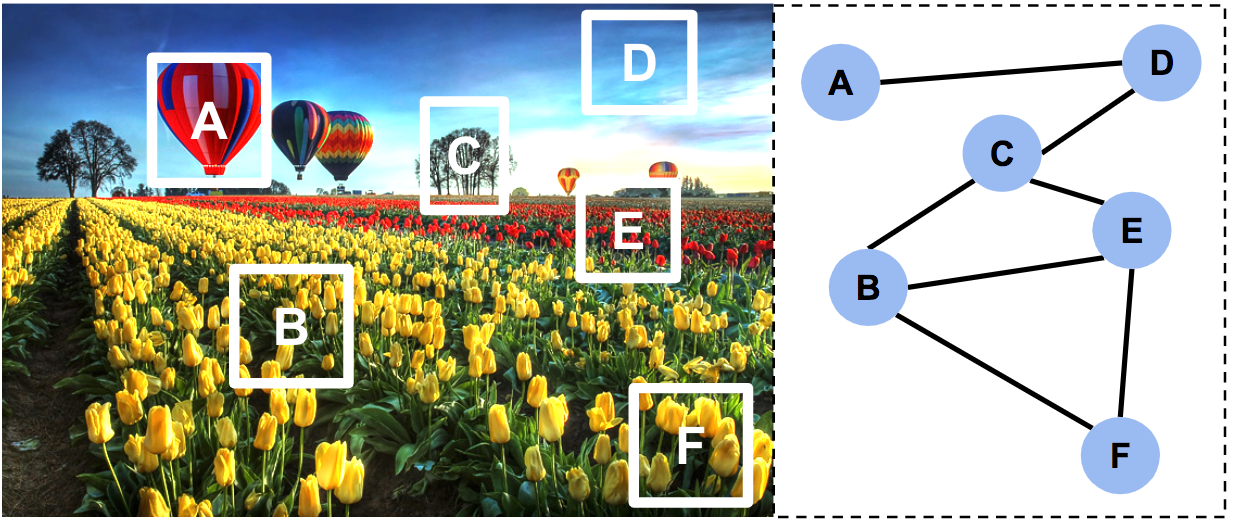}
\end{center}
\vspace{-4mm}
\caption{We represent an image as region composition graph and apply graph convolution for learning aesthetics in the image. To ease the illustration, this figure only shows 7 local regions in the image. In our method, however, an image is partitioned into local regions by a spatial grid and a graph is built over all local regions.}
\vspace{-5mm}
\label{fig:motivation}
\end{figure}

Despite the promising results, these existing methods are typically adapted from the classic image classification networks such as AlexNet~\cite{Krizhevsky:NIPS12},  VGGNet~\cite{Simonyan:ICLR15} and so on, which are not tailored to image aesthetics assessment task and thus not able to model the unique properties related to image aesthetics. Among these properties, the composition information of visual elements in an image plays a crucial role in assessing image aesthetics. In the visual arts, the visual elements in an image never stand alone but rather are mutually dependent on each other and collectively manifest the aesthetics property of the whole image. As illustrated in Figure~\ref{fig:motivation}, the local regions corresponding to ``blue sky'', ``hot balloon'' and ``gorgeous flowers'' highlighted in white boxes show great color harmony and spatial layout, making it very confident to categorize this image as a high aesthetics image. Therefore, it is important to design a network architecture that allows us to encode such information and leverage it to boost the performance.  

\begin{figure*}
\begin{center}
\includegraphics[width=1\linewidth]{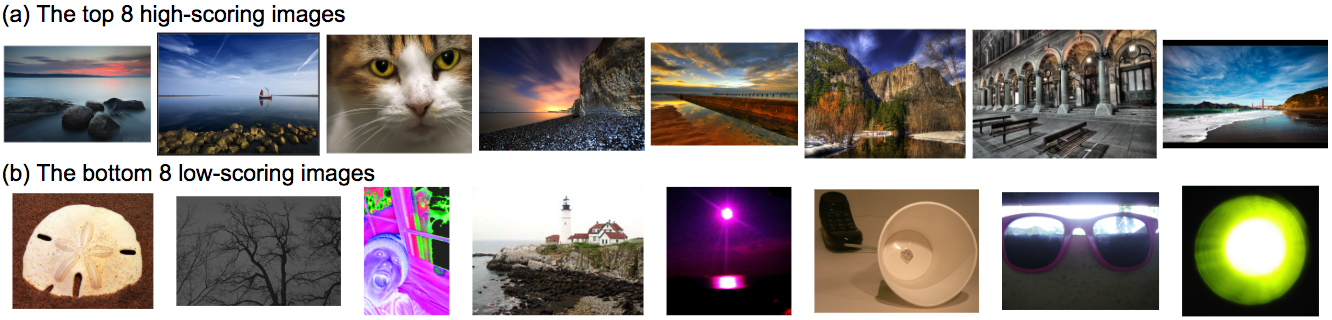}
\end{center}
\vspace{-4mm}  
\caption{The top and bottom scoring images from AVA test set.}
\vspace{-4mm}   
\label{fig:short}
\end{figure*}

In this paper, we propose a composition-aware network, which models the mutual dependencies\footnote{The mutual dependencies explored in this paper are not necessarily the relations between objects, but between local regions.} of visual elements in an image via end-to-end learning. The key idea is straightforward: We densely partition an image into local regions by a spatial grid and compute the features over the regions that can well describe the aesthetics properties of the visual content (i.e., preserving the fine-grained visual details in the image and conveying multi-scale context information surrounding each local region) via an aesthetics feature encoding network backbone (cf. Section~\ref{sec:featureEncodingFCN}). Then the mutual dependencies of image regions can be modeled in a graph-based learning framework. 

As shown in Figure~\ref{fig:motivation}, we represent the input image as a \emph{region composition graph} in which each node represents one region in the image corresponding to one specific spatial position in the image grid. The region nodes are then connected by an edge weighed by the similarity of their features. 
Given the graph representation, we perform reasoning on the graph by applying \emph{Graph Convolution}~\cite{Scarselli:TNN09} operation, in which the activation of each local region is determined by its highly correlated regions. By this learning process, we identify the long range dependencies of local regions in the image and seamlessly leverage them to infer the aesthetics. Since our method uses region graph to model the composition of image, we dub our method \emph{RGNet}, in which the feature learning of local regions and the graph reasoning are performed in an end-to-end manner via a unified network architecture. We will demonstrate experimentally that the proposed method can achieve significant performance gains when evaluated over the benchmark image aesthetics assessment datasets. Figure~\ref{fig:short} shows the top and bottom images in the AVA test set~\cite{Murray:CVPR12} ranked by the estimated aesthetics scores by our model (cf. Section~\ref{sec:experiments}). 

Our main contributions include: (1) An end-to-end image aesthetics assessment network that leverages the composition information of image, (2) a unique aesthetics feature encoding mechanism that can well capture the aesthetics properties of image for visual aesthetics assessment, and (3) state-of-the-art performance with a significant gain in the task of visual aesthetics assessment. 

\begin{figure*}[!htbp]
\begin{center}
\includegraphics[width=.85\linewidth]{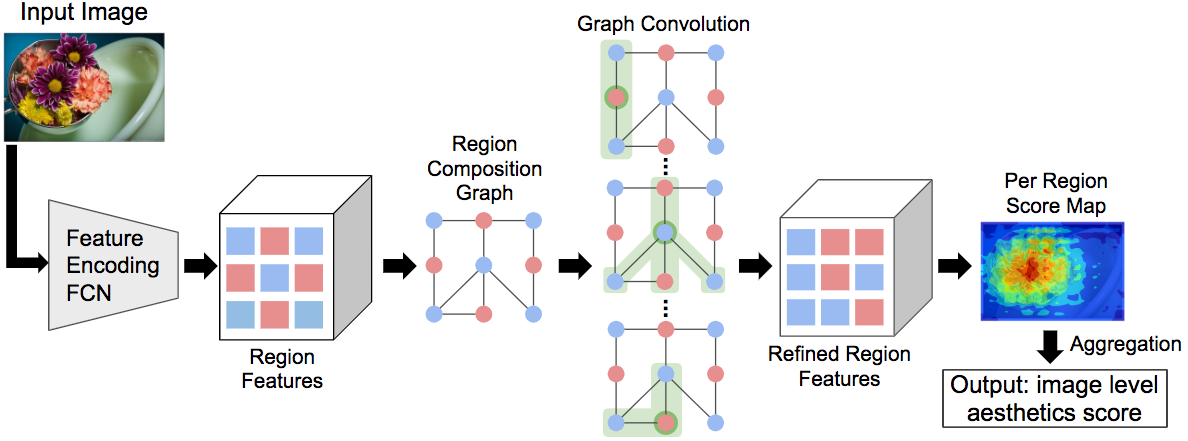}
\end{center}
\vspace{-4mm}
\caption{The \textbf{RGNet} framework for aesthetics prediction. Best viewed in color.} 
\vspace{-4mm}
\label{fig:overview}
\end{figure*}  

\section{Related Work}\label{sec:relatedwork}
\textbf{Aesthetics Prediction with CNN}.  A literature survey on image aesthetics assessment can be found in~\cite{Deng:SP17}, and we will describe the most relevant works here. Recently, there were several efforts to use CNN for image aesthetics prediction. Kong \emph{et al}.~\cite{Kong:ECCV16} formulated aesthetics rating as a ranking problem and trained an AlexNet~\cite{Krizhevsky:NIPS12} inspired Siamese network~\cite{Chopra:CVPR15} architecture to learn the difference of the aesthetics scores of two images. Talebi \emph{et al}.~\cite{Talebi:arXiv17} argued that the human annotated aesthetics ratings in the form of score histograms provided richer information than binary aesthetics labels, and proposed to predict the distribution of aesthetics scores using a CNN model and the squared earth mover's distance loss proposed in~\cite{Hou:NIPS17}. 
Mai \emph{et al}.~\cite{Mai:CVPR16} realized that the image transformations at the data augmentation stage of CNN training may destroy the original image composition and proposed to adopt the \emph{adaptive spatial pooling}~\cite{He:TPAMI15} to directly accept images with original sizes and aspect ratios as input for training CNN. Targeting at the same problem, Ma \emph{et al}.~\cite{Ma:CVPR17} proposed to crop multiple salient patches from the raw image without any transformation, and then build an attribute relation graph over the salient regions to preserve the spatial layout of the image. The features extracted from the salient regions and the vectorized relation graph representation are aggregated into a layout-aware feature vector to represent the image.  Lu \emph{et al}.~\cite{Lu:MM14} proposed a double-column deep convolutional neural network to rate the aesthetics of images, in which one column accepted the whole image as input to model the global view while the other column accepted the randomly-cropped image patch to model the local view. \cite{Lu:ICCV15} argued that using one crop from an image to train an aesthetics CNN model may not capture the information of the entire image and proposed a deep multi-patch aggregation network, which allows the model to be trained with multiple patches generated
from one image. 
Schwarz \emph{et al}.~\cite{Schwarz:WACV18} proposed to encode an image into a high-dimensional feature space resembling visual aesthetics. 
However, all of these methods do not explicitly model the mutual dependencies of visual elements in the image which are an important cue for aesthetics assessment.

\textbf{Visual Dependency modeling}. Modeling the relations of different visual components in visual data has been proven effective in computer vision community. Ma \emph{et al}.~\cite{ Ma:CVPR18} proposed to model higher-order object interactions with attention mechanism for understanding the actions in videos. Wang \emph{et al}.~\cite{Wang:ECCV18} proposed to represent video as a space-time graph which captures temporal dynamics and functional relations between human and object, and then apply graph convolution over the video graph to learn the long range dependencies among the human/object entities in the video. \cite{Wang:CVPR18} proposed a non-local operation for capturing the long-range dependencies among visual elements, and achieved the state-of-the-art results on various computer vision tasks. In image segmentation, modeling the contextual dependency of the local segments with \emph{Condition Random Field} (CRF)~\cite{Krahenbuhl:NIPS11} has become an inevitable step to achieve good performance. Methodologically, our method is closely related to the relation reasoning networks~\cite{Chang:NeurIPS18, Scarselli:TNN09, Santoro:NIPS17} in machine learning community, which were originally proposed to deal with structured data such as texts and speeches. In particular, we are motivated by \emph{Graph Convolution Network}~\cite{Scarselli:TNN09} due to its recent success in computer vision community~\cite{Verma:CVPR18, Wang:ECCV18}. We adopt graph convolution operation as the region dependency modeling mechanism in our aesthetics model, leading to the state-of-the-art results. 

\section{Aesthetics Prediction with RGNet}
Our proposed RGNet aims to represent an image as a graph of local regions and perform reasoning over the graph for aesthetics prediction using an CNN trained end-to-end. Figure~\ref{fig:overview} shows an overview of our model. It takes as input an image and forwards it to a \emph{Fully Convolutional Network} (FCN)~\cite{Shelhamer:PAMI16} style image aesthetics feature encoder dedicated to produce task-tailored features for aesthetics assessment. The output of the FCN is a $3$D feature map in which each spatial location represents one local region in the image and the spatial configuration of all spatial locations in the feature map characterizes the spatial layout of different visual elements in the image. Based on the features of local regions, we construct a \emph{region composition graph} in which each node is a local region and any two nodes are connected by an edge weighted by their feature similarity. With the graph representation, we apply the Graph Convolution operations to perform reasoning. The output of graph convolution is in the same dimension as the input feature map but the mutual dependencies of highly correlated local regions are naturally embedded. Finally, based on the refined feature representation, we predict aesthetics scores for each of the spatial locations and aggregate them to the image level aesthetics score.  

In what follows, we introduce the details of each component of our model. We first describe the aesthetics feature encoding FCN in Section~\ref{sec:featureEncodingFCN}. We then describe the process of applying graph convolution to model the mutual dependencies of different visual elements in Section~\ref{sec:graphConv}. Finally, we discuss the implementation details in Section~\ref{sec:impdetails}.

\subsection{Aesthetics Feature Encoding FCN}\label{sec:featureEncodingFCN}
In principle, we can convert any state-of-the-art CNN architecture into an FCN and use it as our feature encoding backbone, yet choosing the network architecture properly is critical for achieving good performance on aesthetics assessment. In particular, \emph{two special considerations} must be taken into account to ensure that the encoded feature map conveys the aesthetics properties of image. The \emph{first} is to preserve the fine-grained visual details in the image (e.g., the makeup on face) which correspond to the low level features from the shallow layers of a CNN. Such information is crucial for assessing the aesthetics of image and should be fully leverage in model design. To achieve this, we propose to adopt DenseNet~\cite{Huang:CVPR17} as the backbone of our FCN feature encoder. DenseNet uses dense connections to feed the output of each convolution layer to all unvisited layers ahead. In this way, the low level features can be maximally integrated with the semantic features output at the end of the network, and serve as powerful features for learning aesthetics\footnote{As shown in Section~\ref{subsec:ablation}, even a less powerful densely-connected network backbone leads to better performance in asethtics assessment than a more powerful but non densely-connected counterpart.}.  Following the common practice~\cite{Chen:ICLR15},  we convert the DenseNet-121 into an FCN by removing the last two pooling layers and using \emph{atrous convolution}~\cite{Papandreou:CVPR15} to make the pre-trained weights for the convolution layers after the removed layers reusable, in which the dilation rates of the convolution layers after the two removed pooling layers are set as $2$ and $4$. In this way, the converted  DensetNet-121 FCN outputs a feature map of $1/8$ input image resolution.  

\begin{figure}
\begin{center}
\includegraphics[width=\linewidth]{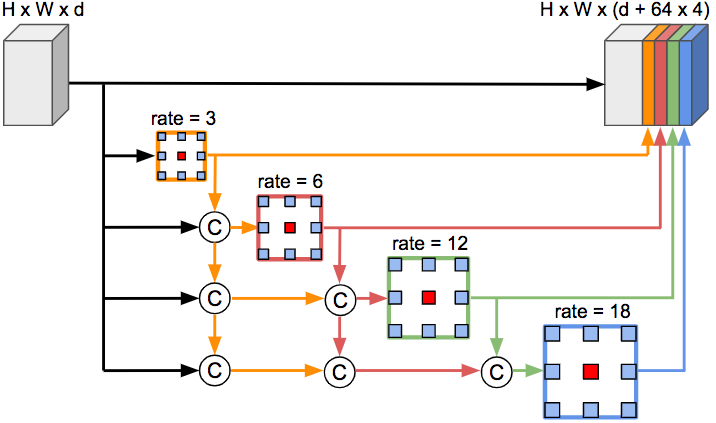}
\end{center}
\vspace{-3mm}
\caption{The block of DenseASPP used in RGNet, where ``C'' denotes channel wise feature concatenation. Best viewed in color.}
\label{fig:FCN}
\vspace{-3mm}
\end{figure}

The \emph{second} consideration is to make the feature map encode multi-scale information in order to convey the diverse range of context in the image. This is important for aesthetics assessment since many photographic images contain object/scene of different scales. Incorporating multi-scale context would help calibrate the scale variances of various visual elements in an image and results in more robust feature representation. To achieve this, \emph{Atrous Spatial Pyramid Pooling} (\emph{ASPP})~\cite{Chen:TPAMI18} is proposed to concatenate feature maps generated by atrous convolution with different dilation rates so that the neurons in the output feature map contain multiple receptive field sizes which encode the multiscale information. In this paper, we adopt the recently proposed \emph{DenseASPP}~\cite{Yang:CVPR18}, which connects a set of atrous convolution layers in a dense way as in DenseNet~\cite{Huang:CVPR17}, to generate such context feature. As illustrated in Figure~\ref{fig:FCN}, given a feature map of dimensions $H\times W\times d$ output from dilated DensetNet-121 where $H \times W$ represents the spatial dimensions and $d$ represents the channel number, we apply $4$ atrous convolution layers with dilation rate of $\{3, 6, 12, 18\}$ in a cascade fashion, each of which produces a feature map of dimensions $H \times W \times 64$. We concatenate these feature maps with the input feature map and end up with a feature map of dimensions $H \times W \times d'$ where $d' = d + 4 \times 64$. 

\subsection{Reasoning Over Region Composition Graph}\label{sec:graphConv}
With the feature map conveying aesthetics properties of the image, we construct a region composition graph over the local image regions in the feature space. In the graph, each node represents a local region, and we connect each pair of nodes with an edge weighed by their similarity. Mathematically, given the FCN feature map of dimensions $H \times W\times d'$, we stack the feature vectors on the individual spatial locations into a matrix $\mathbf{X} = [\mathbf{x}_1, \mathbf{x}_2, \ldots, \mathbf{x}_N]\in\mathbb{R}^{N\times d'}$, where $N=H\times W$ denotes the total number of feature vectors, and each $\mathbf{x}_i \in \mathbb{R}^{d'}$, $i = 1, 2, \ldots, N$ denotes the feature representation of one local region in the image. Then the pairwise similarity function $s(\mathbf{x}_i, \mathbf{x}_j)$ between every two local regions can be defined as their inner product,
\begin{equation}
s(\mathbf{x}_i, \mathbf{x}_j) = \phi(\mathbf{x}_i)^{\top}\theta({\mathbf{x}_j}),
\label{eq:simfunc}
\end{equation}
where $\phi(\mathbf{x}_i) = \mathbf{A}\mathbf{x}_i$ and $\theta(\mathbf{x}_j) = \mathbf{B}\mathbf{x}_j$ are two linear transformations applied on the feature vectors~\cite{Wang:ECCV18, Wang:CVPR18}. The parameters $\mathbf{A}\in \mathbb{R}^{d'\times d'}$ and $\mathbf{B}\in \mathbb{R}^{d' \times d'}$ are weight matrices that can be optimized via back propagation. As shown in~\cite{Wang:ECCV18, Wang:CVPR18}, such linear transformations would increase the generalization ability of learned dependencies.  

After computing the pairwise similarity matrix $\mathbf{S}\in\mathbb{R}^{N\times N}$ with Eq.~(\ref{eq:simfunc}), we perform normalization on each row of the matrix so that the sum of all edge weights connected to one node is $1$. Following~\cite{Wang:CVPR18}, we adopt $\mathtt{softmax}$ function for the normalization:
\begin{equation}
\mathbf{S}_{ij} =\frac{\exp(s(\mathbf{x}_i, \mathbf{x}_j))}{\sum_{k=1}^{N} \exp(s(\mathbf{x}_i, \mathbf{x}_k))},
\label{eq:simmat}
\end{equation}
where $\mathrm{exp}(\cdot)$ denotes the exponential function. The normalized matrix $\mathbf{S}$ is taken as the adjacent matrix representing the relations between the nodes, which characterizes the mutual dependencies of local regions in the image. 

After the region composition graph is constructed, we perform reasoning on the graph by applying \emph{graph convolution}~\cite{Scarselli:TNN09}. Different from the conventional convolution which operates on a local regular grid, graph convolution allows us to compute the response of a node based on its neighbors specified by the graph structure (Figure~\ref{fig:overview}). Therefore, performing graph convolution over the feature map output by our feature encoding FCN is equal to perform message passing across local regions in the image. The outputs of graph convolution are the enhanced feature representations for each local region, where the mutual dependencies of regions are naturally encoded. Formally, we can define the graph convolution as, 
\begin{equation}
\mathbf{Z} = \mathbf{S}\mathbf{X}\mathbf{W},
\end{equation}
where $\mathbf{W}\in \mathbb{R} ^{d' \times d'}$ is the weight matrix for one graph convolution layer and $\mathbf{Z} \in \mathbb{R}^{N\times d'}$ is the output feature from the layer. In this work, we stack three graph convolution layers as the region dependency learning module. After each layer of graph convolution, we apply $\mathtt{ReLU}$ activation function on the output feature map $\mathbf{Z}$. Denote by $\mathbf{G}\in\mathbb{R}^{N\times d'}$ the final feature map after the stack of graph convolutions, we reshape it back to a  feature map of dimensions $H\times W \times d'$ for applying the classification head.  

\textbf{Aesthetics Classification Head}. The updated feature map is then forwarded to the network \emph{head} for inferring the aesthetics score of image. To this end, we use a $1 \times 1 \times d'$ small kernel followed by $\mathtt{softmax}$ function to produce one score $Y_{ij}^{c}\in [0, 1]$ at each of the $H\times W$ spatial locations on the feature map and for each of the aesthetics classes $c\in \{0,1\}$, where $0$ denotes the class of ``\emph{low easthetics}'' and $1$ denotes that of ``\emph{high aesthetics}'' (Figure~\ref{fig:overview}). Given that we have to predict the aesthetics label at the image level, we need to aggregate these region-level scores into a single image-level aesthetics score. Following~\cite{Pinheiro:CVPR15}, we choose a smooth convex function called $\textit{Log-Sum-Exp}$ (LSE) as our aggregation function:
\begin{equation}
y^{c} = \frac{1}{r}\log \bigg(\frac{1}{HW}\sum_{i,j}\exp(rY^{c}_{ij}) \bigg),
\label{eq:agg}
\end{equation}
where $r$ is a hyper-parameter controlling the smoothness of the approximation: high $r$ values imply having an effective similar to the max pooling while very low values will have an effect similar to the score averaging. The advantage of this aggregation function is that local regions with similar scores will have a similar weight in the training procedure with $r$ controlling the notion of similarity. In this work, we use $r=4$ according to the validation performance. The output $y^{c}\in\mathbb{R}$ denotes the image level aesthetics score aggregated over the local regions. We further convert these image-level scores into class conditional probabilities by applying a $\mathtt{softmax}$ function:  
\begin{equation}
p(c| I, \mathbf{w}) = \frac{\exp(y^c)}{\sum_{k\in\{0,1\}}\exp(y^k)},
\label{eq:aggsf}
\end{equation}
where $I$ is the input image and $\mathbf{w}$ denotes all trainable parameters of our network architecture.

\subsection{Implementation Details}\label{sec:impdetails}
\textbf{Training}. Our method is implemented with PyTorch library~\cite{Paszke:NIPS17}. The DenseNet-121 network~\cite{Huang:CVPR17} used as the backbone of our aesthetics feature encoding FCN is pre-trained on ImageNet~\cite{Russakovsky:IJCV15}. After assembling our RGNet architecture, we fine-tune the entire network end-to-end with images in the domain of aesthetics prediction. To train the network, we use the  \emph{binary cross entropy} loss as the training objective function. During training, the input RGB images are first resized to the resolution of $300 \times 300$. With the input image size, our aesthetics feature encoding FCN outputs a feature map of spatial size $37\times 37$. To avoid over-fitting, we adopt common data augmentations on the resized images as preprecessing, including random flipping training images horizontally with the probability of $0.5$, randomly scaling images in the range of $[1.05, 1.25]$ times of the input image size followed by randomly cropping $300 \times 300$ image patches. 

We train the model on a $4$-GPU machine with a min-batch size of $32$ images. We choose Adam optimizer~\cite{Kingma:ICLR15} and train our model for $80$ epochs in total, starting with a learning rate of $10^{-4}$ and reducing it by multiplying the initial learning rate with $\mathrm{(1-epoch/maxEpochs)^{0.9}}$, where $\mathrm{epoch}$ is the value of current epoch and $\mathrm{maxEpochs}$ is the total number of epochs for training. During training, the weight decay is set as $10^{-5}$ and batch normalization~\cite{Ioffe:ICML15} is used before each weight layer to ease the training and make the concatenated features  from different layers comparable. We adopt the method in~\cite{He:ICCV15} to initialize the weights of the newly added layers in the network. 

\textbf{Inference}. At test time, we perform fully convolutional inference to get the aesthetics score on the test image. Specifically, we pass each image through the network to get the aesthetics score of each local region in the image and then use Eq.~(\ref{eq:agg}) and Eq.~(\ref{eq:aggsf}) to aggregate them into the image-level aesthetics score. The final classification label is decided by comparing the scores between positive and negative aesthetics classes. 

\section{Experiments}\label{sec:experiments}
In this section, we evaluate the effectiveness of the proposed RGNet on the image aesthetics assessment task. We first introduce the datasets and then perform comprehensive ablation experiments along with a thorough comparison of RGNet to the state of the art. 

\subsection{Datasets}
\noindent \textbf{Aesthetic Visual Analysis (AVA) dataset}~\cite{Murray:CVPR12}. To our best knowledge, AVA is the largest publicly available image aesthetics assessment dataset. Due to its representativeness, some recent CNN based aesthetics assessment works conduct experiments solely on this dataset~\cite{Lu:MM14, Murray:arxiv2017, Mai:CVPR16, Ma:CVPR17}. It contains around $250, 000$ images downloaded from \emph{DPChallenge}~\cite{DP:URL}, an online community for amateur photographers, in which the users rate and comment on each other's photographs in response to some digital photography contests. In AVA dataset, each image is scored by an average of $210$ users and the image ratings range from $1$ to $10$ with $10$ being the highest aesthetic score of an image. To ensure fair comparison, we follow the same training/test data partition of the AVA dataset as the previous work~\cite{Lu:MM14, Lu:ICCV15,  Murray:arxiv2017, Mai:CVPR16}, in which there are around $230,000$ images for training and $20,000$ images for testing. From the training set, we further hold out $2,000$ images as the validation set for validating model hyper-parameters. We also use the same mechanism to assign a binary aesthetics label to each image in the dataset. Specifically, images with mean ratings smaller than $5$ are labeled as low-aesthetics images and those with mean ratings larger than or equal to $5$ are labeled as high-aesthetics images. We report the binary classification accuracy, the standard AVA evaluation metric for all the experiments. We also report Average Precision (AP) on AVA to evaluate how well the model is able to rank images based on the estimated aesthetics scores. 

\noindent \textbf{Aesthetics and Attribute Database (AADB)}~\cite{Kong:ECCV16}. It contains $10,000$ photographic images downloaded from \emph{Flickr}~\cite{Flickr:URL}, each of which has a real valued aesthetics score aggregated across five aesthetics ratings provided by five different individual raters. The dataset is split into training ($8,500$), validation ($500$) and test ($1,000$)
sets, in which we train our model on the training set and tune the hyperparamters on the validation set.  Since the images have real valued score as ground truth, we train a regression RGNet by replacing the \emph{cross entropy} loss with \emph{mean square error} loss.  We report the Spearman's rank correlation coefficient $\rho$,  the standard evaluation metric on AADB dataset, to measure the model performance. 

\begin{table}
\begin{center}
\begin{tabular}{l|c|c|c}
Method & Input Image Size & Acc. ($\%$) & AP\\
\Xhline{3\arrayrulewidth}
FC-CNN & $224\times 224$ &$80.45$ & $0.9005$ \\
FCN & $300\times 300$ & $81.43$ & $0.9112$\\
FCN-A & $300\times 300$ & $82.11$ & $0.9262$\\
FCN-G & $300\times 300$ & $82.33$ & $0.9277$\\
\hline
FCN-C-C & $300\times 300$ & $80.52$  & $0.9038$\\
\textbf{FCN-A-G} & $300\times 300$ & $\textbf{83.59}$ & $\textbf{0.9462}$\\
\end{tabular}
\end{center}
\vspace{-2mm}
\caption{Results on AVA test set for different variants of the proposed model, comparing different architecture design choices. Since FC-CNN is fine-tuned on the pretrained fully connected DenseNet-121, its input size has to be the same size of $224\times 224$.} 
\label{tab:components}
\vspace{-4mm}
\end{table}

\subsection{Ablation Studies}\label{subsec:ablation}
\textbf{Model Component Analysis}. Our RGNet model is composed of different components that serve different purposes. In this section, we will run a number of experiments to study how the individual components contribute to the overall performance quantitatively. We conduct analysis over AVA dataset considering it is a large scale dataset and compare the following $6$ variants of the model: \vspace{-2mm}

\begin{itemize}[leftmargin=*]
\item \emph{Fully Connected CNN} (FC-CNN): This model is identical to the fully connected DenseNet-121~\cite{Huang:CVPR17}, in which the input image is encoded into a $\mathrm{1D}$ feature vector and then fed into the fully connected layers for classification. \vspace{-6mm}
 \item \emph{FCN}: This model is a truncated version of RGNet, in which the ASPP and Graph Convolution modules are removed.\vspace{-2mm}
 \item \emph{FCN-ASPP} (FCN-A): This model further adds the ASPP module of RGNet into the \emph{FCN} model specified above.\vspace{-2mm}
 \item \emph{FCN-Graph} (FCN-G): As the name indicates, this model contains the FCN and graph convolution modules, but the ASPP module is excluded.\vspace{-2mm}
 \item \emph{FCN-ASPP-Graph} (FCN-A-G): This is our proposed RGNet model including every component. \vspace{-2mm}
 \item \emph{FCN-Conv-Conv} (FCN-C-C): This baseline model is added in order to verify that the improvement of RGNet is \emph{not} just because it adds depth to the network backbone. Specifically, we replace both ASPP and graph convolution blocks with the same number of convolution layers, and compare the model with FCN-A-G. 
\end{itemize}
\vspace{-1.5mm}


The convolution layers of each of the $6$ models are pre-trained on ImageNet~\cite{Russakovsky:IJCV15}, based on which we fine-tune the entire network end-to-end. For fair comparison, we adopt the same experiment settings and implementation details across all of the models. The performance of these models is shown in Table~\ref{tab:components}.  From the results, we have the following observations: (1) FCN performs better than FC-CNN. This is due to the fact that the feature map of FCN has a spatial resolution of $37\times 37$, and thus is able to comprehensively capture the visual elements and their spatial relations. On the contrary, FC-CNN completely loses the composition information in the image when it encodes the image into a $\mathrm{1D}$ feature vector. (2) Both FCN-A and FCN-G outperform FCN. This indicates that modeling multiscale context information (ASPP) or mutual dependencies among local image regions (Graph Convolution) benefits aesthetics prediction. (3) The FCN-A-G (i.e., RGNet) further beats FCN-A and FCN-G and achieves the best performance over all models in comparison. This may attribute to the fact that the graph convolution that follows ASPP obtains additional context information, making the learned mutual dependencies more accurate. The success of FCN-A-G confirms the value of each individual component in our network architecture design. (4) Although FCN-C-C has similar depth as RGNet, its performance is much worse. This shows that improvement of RGNet is not just because it adds depth. 

\begin{table}
\begin{center}
\begin{tabular}{l|c|c}
Method & Acc. ($\%$) & AP \\
\Xhline{3\arrayrulewidth}
VGG-16 FCN & $76.32$ & $0.8806$\\
ResNet-101 FCN & $80.44$ & $0.9013$\\
\hline
RGNet (VGG-16 FCN) & $78.21$ & $0.8905$\\
RGNet (ResNet-101 FCN) & $82.53$ & $0.9372$\\
RGNet (DenseNet-121 FCN) &  $\textbf{83.59}$ & $\textbf{0.9462}$\\
\end{tabular}
\end{center}
\vspace{-3mm}
\caption{Results on the AVA test set for the proposed method with different network backbone architectures. The input image size for all models is $300\times 300$.} 
\label{tab:architecture}
\vspace{-5mm}
\end{table}

\textbf{Backbone Architecture}. To demonstrate that the choice of DenseNet architecture as our feature encoding backbone helps generate task-tailored features for aesthetics assessment, we instantiate RGNet with different convolutional backbone architectures for the feature encoding FCN and compare their performance. To this end, we evaluate VGG-16~\cite{Simonyan:ICLR15} and ResNet-101~\cite{He:CVPR16} and reconfigure each of them into an FCN. The recongaration is done following the strategy in~\cite{Chen:TPAMI18}. Specifically, given the VGG-16 architecture, we first convert the fully-connected layers into convolutional ones to make it an FCN. To ensure the output feature map size the same as RGNet (i.e., $1/8$ of the input image resolution), we skip the subsampling after the last two max-pooling layers in the network, and set the dilation rates of the convolution layers that follow them to be $2$ and $4$.  As for ResNet-101, we do similar conversion in that we remove the last classification layer and modify the stride of the last two convolution layers from $2$ to $1$, followed by setting the dilation rates of these convolution layers to be $2$ and $4$, making the resolution of the output feature map $1/8$ times the input image size. It is worth noting that ResNet-101 is a more powerful network architecture with around $5\small{\times}$ parameters than DenseNet-121, and its performance on the ImageNet benchmark is better than the latter~\cite{Huang:CVPR17}.    

\begin{figure}
\begin{center}
\includegraphics[width=0.95\linewidth]{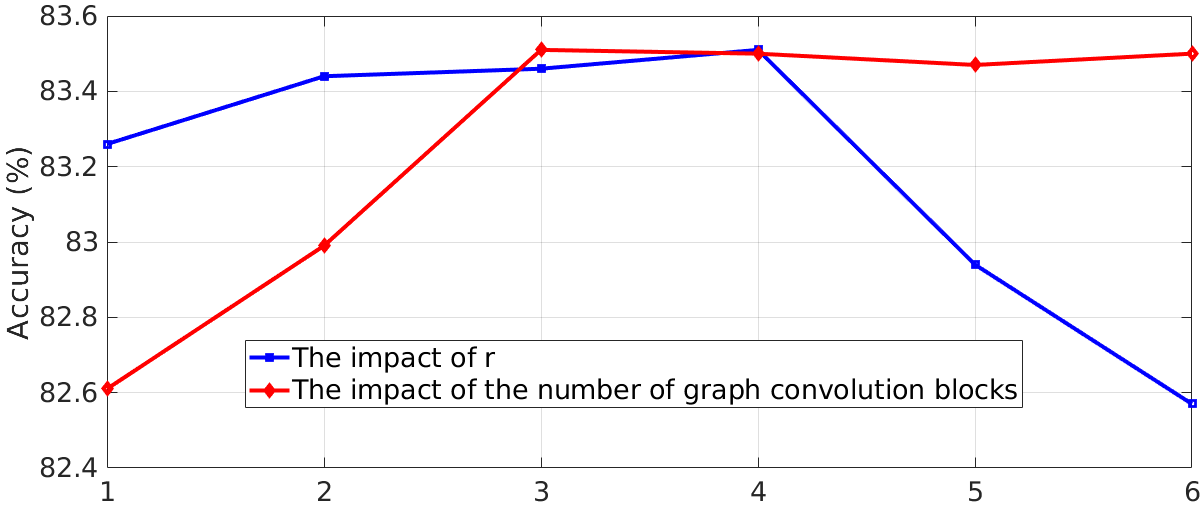}
\end{center}
\vspace{-3mm}
\caption{Model Performance on the validation set by varying the value of $r$ and the number of graph convolution blocks.}
\label{fig:hyperparameter}
\vspace{-6mm}
\end{figure}

Table~\ref{tab:architecture} shows the results of RGNet with different backbones (rows 4-6). From the results, we can see that RGNet benefits from densely connected network and DenseNet-121 backbone clearly outperforms VGG-16 and ResNet-101 counterparts. This complies with our assumption that fine-grained low-level features conveyed by the dense connection of DenseNet well preserve the aesthetics properties of image and therefore benefit the task of aesthetics assessment. On the contrary, the non densely-connected network backbones such as ResNet-101, even more powerful in the general image classification task, may not be able to generate task-tailored features for aesthetics assessment.     

It is interesting to see how the above ``degraded'' versions of RGNet based on non densely-connected network backbones perform comparing to the state of the art. To do this, we can compare the performance of the above two models in Table~\ref{tab:architecture} to the state-of-the-art results in Table~\ref{tab:AVASoA}.  Note that most of the methods in Table~\ref{tab:AVASoA} leverages additional information such as scene context or aesthetics attributes. Without any of such information, the performance of our RGNet based on VGG-16 is among the top 4 ($78.21\%$). When the more powerful ResNet-101 backbone is applied, our method outperforms APM~\cite{Murray:arxiv2017} which is also based on ResNet-101 ($82.53\%$ \emph{vs} $80.30\%$) and the state-of-the-art result in literature ($82.53\%$ \emph{vs} $82.50\%$). This validates the superiority of our RGNet over the existing methods.   

We also present the results of VGG-16 FCN and  ResNet-101 FCN (Table~\ref{tab:architecture} rows 2-3) as baselines to show the performance improvement introduced by the proposed RGNet when different network backbones are applied as feature encoding FCN. As seen, adding the additional modules in RGNet, including ASPP and graph convolution, would consistently boost the performance of the two baseline models by $1.89\%$ and $2.09\%$ respectively. This confirms the generalization ability of the proposed method.  

\textbf{Model Hyperparameters}. In training our model, the smoothness parameter $r$ in Eq~(\ref{eq:agg}) has to be chosen to get good aggregation results. Figure~\ref{fig:hyperparameter} plots the values of classification accuracy on the validation set as a function of different values of $r$. As seen, the model achieves the best performance when $r=4$, which is close to the value suggested in~\cite{Pinheiro:CVPR15}. As $r$ grows, the aggregation function deemphasizes regions with low scores, and eventually decreases the performance due to the information loss. Therefore, we fix $r=4$ for training the RGNet model. Our aggregation function achieves $0.3\%$ performance gains in terms of classification accuracy than \emph{Average Pooling} on AVA test set. 

The number of graph convolution blocks is the other parameter of our model. Figure~\ref{fig:hyperparameter} shows the results of our model with different number of graph convolution blocks on the validation set of our experiment. It shows that the performance remains stable after $3$ graph convolution blocks are applied. Considering that adding more graph convolution blocks would increase the number of parameters and FLOPs, we choose to use $3$ blocks of graph convolution to achieve a good tradeoff between accuracy and model size. 

\begin{table}
\begin{center}
\begin{tabular}{l|c|c}
Layers with different dilation rates & Acc. ($\%$)  & AP  \\
\Xhline{3\arrayrulewidth}
(3, 6) & $82.72$ & $0.9338$\\
(6, 12) & $82.67$ & $0.9327$\\
(3, 6, 12) & $83.16$ & $0.9410$\\
(3, 6, 12, 18, 24) & $83.49$ & $0.9469$ \\
\hline
(3, 6, 12, 18) & $\textbf{83.51}$ & $\textbf{0.9485}$ \\
\end{tabular}
\end{center}
\vspace{-2mm}
\caption{Performance of RGNet with different number of atrous convolution layers employed by ASPP on the validation set.} 
\label{tab:aspp}
\vspace{-4mm}
\end{table}

Another parameter is the number of the atrous convolutional layers employed in ASPP. To find the best configuration, we use different number of atrous convolution layers with different dilation rates, and see how they influence the performance of RGNet. The results are evaluated on the validation set of our experiments, and illustrated in Table~\ref{tab:aspp}.  We observe that using $4$ atrous convolution layers with dilation rates of $\{3, 6, 12,18\}$ for ASPP achieves the best performance. Therefore, we adopt this configuration in the ASPP module of our model.  

\begin{table}
\begin{center}
\begin{tabular}{l|c}
Method & Acc. ($\%$) \\
\Xhline{3\arrayrulewidth}
Hand-crafted feature + SVM~\cite{Murray:CVPR12} & $66.70$ \\
\hline
RDCNN~\cite{Lu:MM14} & $74.46$ \\
DMA-Net-ImgFu~\cite{Lu:ICCV15} & $75.41$ \\
Aesthetics-Encoding~\cite{Schwarz:WACV18} & $75.83$\\
MT-CNN~\cite{Kao:TIP17} & $76.58$ \\
BDN~\cite{Wang:IJCNN17} & $76.80$  \\
Regression+Rank+Att+Cont~\cite{Kong:ECCV16} & $77.33$ \\
MNA-CNN-Scene~\cite{Mai:CVPR16} & $77.40$ \\
APM~\cite{Murray:arxiv2017} & $80.30$ \\
NIMA~\cite{Talebi:arXiv17} & $81.51$ \\
A-Lamp~\cite{Ma:CVPR17} & $82.50$ \\
MPA~\cite{Sheng:MM18} & $83.03$\\
\hline
\textbf{RGNet (our)} & $\textbf{83.59}$ \\
\end{tabular}
\vspace{-3mm}
\end{center}
\caption{Comparison to the state-of-the-art aesthetics assessment performance on the standard AVA test set.} 
\vspace{-6mm}
\label{tab:AVASoA}
\end{table}

\subsection{Comparing to state-of-the-art methods} 
We compare the proposed RGNet to the state-of-the-art aesthetics prediction models on both AVA and AADB datasets and the results can be found in Table~\ref{tab:AVASoA} and Table~\ref{tab:AADBSoA} respectively. All these results are based on the same training/test split of the dataset and are directly quoted from their original papers. As seen, our method outperforms the state of the art by wide margins. In particular, it is interesting to highlight the comparison between our method and the following two baseline methods listed in Table~\ref{tab:AVASoA} to better understand our contributions. Specifically, the baseline MNA-CNN-Scene~\cite{Mai:CVPR16} noticed that the input image pre-processing such as cropping, scaling or padding in CNN aesthetics model training damages image composition, and proposed to adopt \emph{Spatial Pyramid Pooling} (SPP)~\cite{He:TPAMI15} to directly handle input images with original sizes and aspect ratios. Although this method is called \emph{composition-preserving} aesthetics model, it does not explicitly represent the visual elements in the image or model their mutual dependencies as our method does. Our RGNet model shows $6.19\%$ absolute improvement on the classification accuracy. The other baseline method A-Lamp~\cite{Ma:CVPR17} is an image layout aware method that aggregates the vectorized representation of a spatial relation graph built over salient regions of an image with the features of individual patches as the global image feature to learn aesthetics. It only focuses on spatial layout of image and does not device a mechanism to model the mutual dependencies of visual elements at the feature level as our method does. Without bells and whistles, our RGNet model outperforms A-Lamp by $1.09\%$. We expect many such improvements as in the above two baselines to be applicable to our method.  Table~\ref{tab:AADBSoA} lists the performance of different methods on AADB dataset.  Again, our method beats the best result by $3.22\%$. 
\begin{table}
\begin{center}
\begin{tabular}{l|c}
Method & $\rho$ \\
\Xhline{3\arrayrulewidth}
Finetune-Conf~\cite{Kong:ECCV16} & $0.5923$ \\
Regression~\cite{Kong:ECCV16} & $0.6239$ \\
Regression+Rank~\cite{Kong:ECCV16} & $0.6515$ \\
Regression+Rank+Att+Cont~\cite{Kong:ECCV16} & $0.6782$\\
\hline
\textbf{RGNet (our)} & $\textbf{0.7104}$ \\
\end{tabular}
\vspace{-2mm}
\end{center}
\caption{Comparison to the state-of-the-art aesthetics assessment performance on the standard AADB test set.} 
\vspace{-5mm}
\label{tab:AADBSoA}
\end{table}

\begin{figure}
\begin{center}
\includegraphics[width=\linewidth]{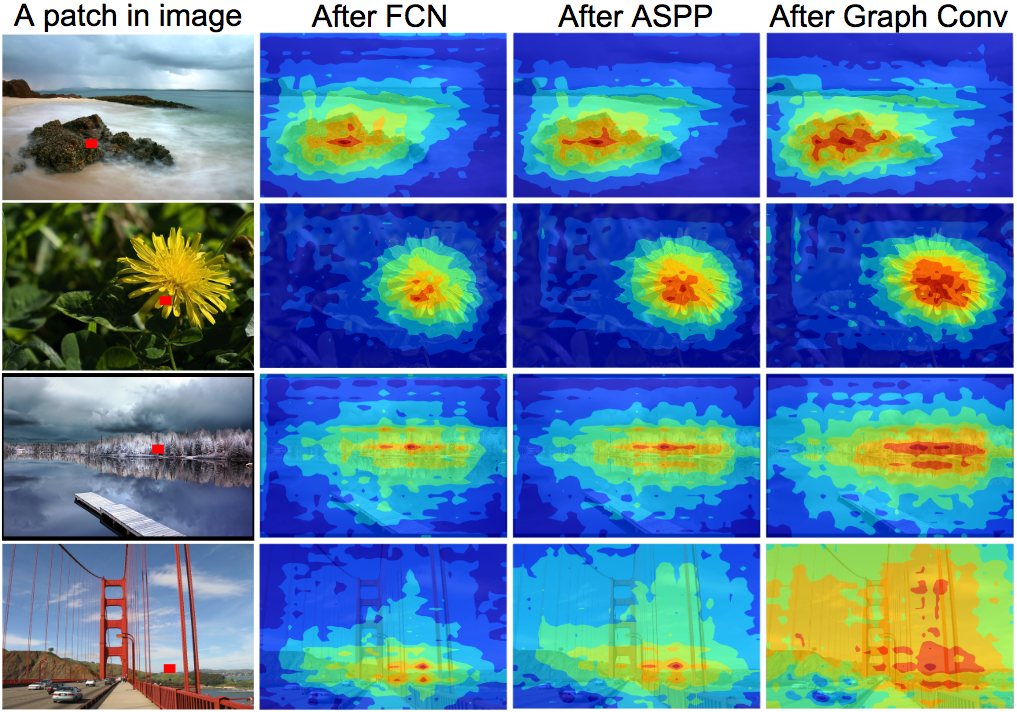}
\end{center}
\vspace{-3mm}
\caption{Feature similarities of all regions to a specified region marked by a red rectangle in the image. Hotter color indicates higher similarity in feature space. Best viewed in color. }
\vspace{-6mm}
\label{fig:heatmap}
\end{figure}

\subsection{Visualizing the learned region dependency}
The features learned by our RGNet model are able to capture the mutual dependency of local regions in the image. To illustrate this, we perform feature level analysis to see how a local region in the images correlates with other regions. Recall that the output of RGNet is a $\mathrm{3D}$ feature map, in which each spatial position of the feature map corresponds to one region, and regions that are highly correlated are likely to have similar features. Therefore, feature similarities between regions are calculated to reveal the learned region dependencies. More specifically, given an image, we first localize a region which has the highest aesthetics score and then compute the feature similarities between the query region and all other regions in the image. Here cosine similarity function is used to measure similarities between features of region. To calculate the similarities, we use features from three different stages of our RGNet network architecture which are respectively features \emph{After FCN}, \emph{After ASPP} and \emph{After Graph Convolution}, and compare how they describe the region dependency differently.  

The results are visualized as heatmap and shown in Figure~\ref{fig:heatmap}. From the results, we can see that the similarity map generated by the features after graph convolution has the largest continuous hot area that covers the major object/scene that is highly correlated with the query patch in the image. For example, the hot area in the first image almost covers the entire rock on which the query image region lies.  On the other hand, the hot areas are much smaller when using features from the stages where graph convolution is not applied. This indicates that the graph convolution further enhances the features from FCN and ASPP by incorporating the long range dependencies of local regions. 

\section{Conclusion}
We have developed RGNet, an end-to-end network architecture for learning image aesthetics from the composition information of local regions in the image. RGNet builds a region graph to represent the visual elements and their spatial layout in the image, and then performs reasoning on the graph to uncover the mutual dependencies of the local regions, leading to the state-of-the-art performance on the benchmark visual aesthetics datasets. As future work, we will adapt the RGNet into a conditional \emph{Generative Adversarial Network}~\cite{Goodfellow:NIPS14} to synthesize high aesthetics images based on the composition information of image.


{
\small
\bibliographystyle{ieee}

}

\end{document}